\documentclass[letterpaper, 10 pt, conference]{ieeeconf}
\usepackage[T1]{fontenc}
\usepackage{CJKutf8}
\usepackage{amsfonts}
\usepackage{amsmath}
\usepackage{siunitx}
\usepackage{booktabs}
\usepackage{multirow}
\usepackage{graphicx}
\usepackage{cite}
\usepackage{balance}

\usepackage[dvipsnames]{xcolor}
\usepackage{pifont}
\newcommand{\cmark}{\textcolor[HTML]{59a14f}{\ding{51}}}%
\newcommand{\xmark}{\textcolor[HTML]{e15759}{\ding{55}}}%

\IEEEoverridecommandlockouts                              

\title{\LARGE \bf
Hierarchical Visual Policy Learning for Long-Horizon Robot Manipulation in Densely Cluttered Scenes
}

\author{Hecheng Wang$^{1}$, Lizhe Qi$^{1*}$, Bin Fang$^{2}$ and Yunquan Sun$^{1}$
\thanks{$^{1}$Academy for Engineering \& Technology, Fudan University, Shanghai, China. 
}%
\thanks{$^{2}$Department of Computer Science and Technology, Tsinghua University, Beijing, China. }%
\thanks{*Correspondence: {\tt\small qilizhe@fudan.edu.cn}}%
}

\begin{document}

\maketitle
\thispagestyle{empty}
\pagestyle{empty}

\begin{abstract}
In this work, we focus on addressing the long-horizon manipulation tasks in densely cluttered scenes. Such tasks require policies to effectively manage severe occlusions among objects and continually produce actions based on visual observations. We propose a vision-based Hierarchical policy for Cluttered-scene Long-horizon Manipulation (HCLM). It employs a high-level policy and three options to select and instantiate three parameterized action primitives: push, pick, and place. We first train the pick and place options by behavior cloning (BC). Subsequently, we use hierarchical reinforcement learning (HRL) to train the high-level policy and push option. During HRL, we propose a Spatially Extended Q-update (SEQ) to augment the updates for the push option and a Two-Stage Update Scheme (TSUS) to alleviate the non-stationary transition problem in updating the high-level policy.
We demonstrate that HCLM significantly outperforms baseline methods in terms of success rate and efficiency in diverse tasks. We also highlight our method's ability to generalize to more cluttered environments with more additional blocks.

\end{abstract}

\section{INTRODUCTION}
In unstructured environments such as homes and offices, robots are expected to help humans perform diverse long-horizon manipulation tasks\cite{BEHAVIOR1K}. The long-horizon property of these tasks significantly increases the task complexity, as policies must continually produce actions based on visual observations over prolonged periods.
To mitigate this challenge, one main approach uses temporal abstraction\cite{option-framwork} to leverage the hierarchical structure of tasks. Specifically, long-horizon tasks can be split into multiple short-horizon subtasks. Those with similar actions can be grouped into a subtask family and addressed using a single action primitive (or skill). While previous primitive-based works\cite{Transporter,CLIPort,MAPLE,RLBC,Romanet,PAC-Romannet} have achieved impressive results under certain conditions, they often assume objects are not occluded\cite{Transporter,CLIPort,MAPLE,RLBC} or require objects' poses as input\cite{MAPLE}, which restricts their application in densely cluttered scenes common in unstructured environments.


Recent works \cite{push_to_grasp1,push_to_grasp2,push_to_grasp3,push_to_grasp4,VPG} have explored the use of push and pick primitives for tasks like object removal in cluttered scenes. However, the placement of objects afterward remained unaddressed. Other works \cite{pick_to_place1,pick_to_place2,pick_to_place3,pick_to_place4,pick_to_place5} investigated the use of pick and place primitives for tasks like object rearrangement in cluttered scenes. However, the necessity to continually remove obstacles by picking and placing reduces the efficiency of these methods. Building on the methods above, several studies\cite{Romanet,PAC-Romannet,Prehensile_and_Non-Prehensile,Selective_Object_Rearrangement,Synergistic-Task-and-Motion-Planning,Towards-Practical-Multi-Object-Manipulation,Tree-Search-based-Task-and-Motion-Planning} have focused on the concurrent use of push, pick, and place primitives to accomplish tasks. However, some of these methods\cite{Synergistic-Task-and-Motion-Planning,Towards-Practical-Multi-Object-Manipulation,Tree-Search-based-Task-and-Motion-Planning} are limited by their assumptions of complete state observability and known object models, while other methods\cite{Romanet,PAC-Romannet,Prehensile_and_Non-Prehensile,Selective_Object_Rearrangement} employ an implicit greedy policy to select primitives rather than explicit planning. 
This prevents the policy from leveraging synergies between primitives, limiting its performance and adaptability.

\begin{figure}[t]
\centering
  \includegraphics[width=\linewidth,scale=1.00]{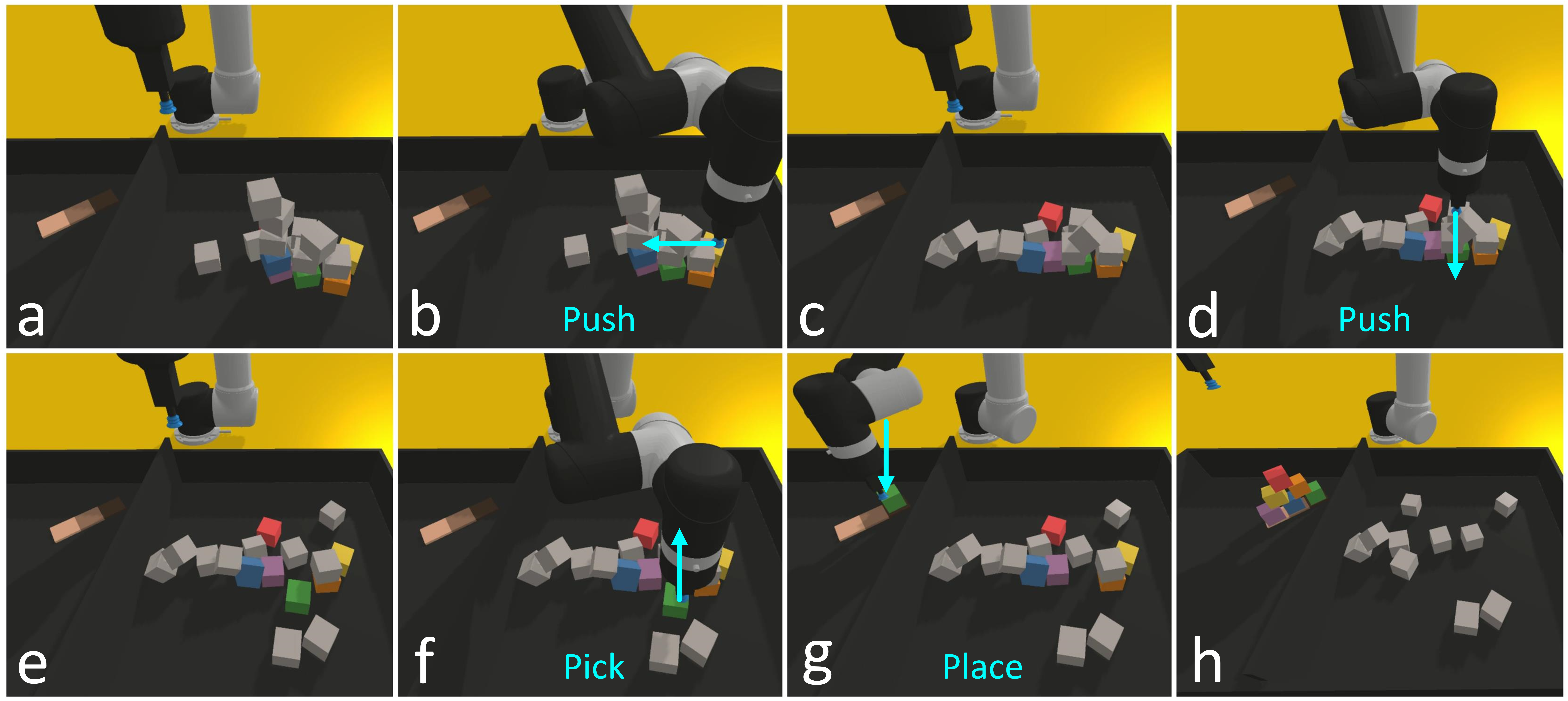}
\vspace{-1.7em} 
  \caption{Example of a trained HCLM policy on the cluttered-stack-block-pyramid task (see Table \ref{tab:Tasks and Attributes}). Initially (a), the target blocks (colored) and additional blocks (grey) are stacked together, representing the clutter commonly observed in real-world scenes. Then (b-g), our policy declutters with two pushes, followed by pick and place. The following steps are skipped, presenting the result in (h).
  }
  \label{fig:Intro}
  \vspace{-2.0em} 
\end{figure}

To effectively perform long-horizon manipulation tasks in densely cluttered scenes, we propose a hierarchical policy harnessing three parameterized action primitives: push, pick, and place. This policy comprises a high-level policy and three options,
each bridging a different primitive. Given the visual observation, the high-level policy outputs a high-level action to select primitive(s) for execution. Meanwhile, the options produce low-level actions to derive parameters for each primitive, which outputs a temporally extended behavior. Specifically, we design a Dual-Level Action Network (DLAN) to model the high-level policy and push option. Inspired by the feature extraction and fusion techniques in CLIPort\cite{CLIPort} and semantic segmentation\cite{two-stream}, we propose a two-stream Transporter to model the pick and place options. We first employ behavior cloning (BC) to train two-stream Transporter to overcome the exploration challenge caused by the sparse pick and place reward. Then, we use hierarchical reinforcement learning (HRL) to train DLAN to avoid repeated collecting of training data.
During HRL, we design a Spatially Extended Q-update (SEQ) to augment the update method for the push option. To alleviate the non-stationary transition problem in HRL, we propose a Two-Stage Update Scheme (TSUS) to modulate updates to the high-level policy. See Fig. \ref{fig:Intro} for an example of our trained policy.

We extend Ravens\cite{Transporter} benchmark to ClutteredRavens, which includes six cluttered-scene manipulation tasks with varying attributes and complexities. Extensive experiments are conducted on ClutteredRavens. Compared to the baselines, HCLM achieves the highest success rate across all tasks and the shortest episode length in four tasks. We also validate the adaptability of HCLM to more cluttered environments with more additional blocks. Ablation studies validate the effectiveness of our proposed modules.


The main contributions of this work include the following: 1) We propose a vision-based Hierarchical Policy for Cluttered-scene Long-horizon Manipulation (HCLM) to accomplish tasks by coordinating push, pick, and place primitives. 2) We propose two heuristic methods, SEQ and TSUS, to augment Q-value updates and mitigate the non-stationary transition problem in HRL. 3) We conduct extensive comparisons with baselines on diverse cluttered-scene long-horizon manipulation tasks, validating the effectiveness of our approach.

\vspace{-0.5em}
\section{RELATED WORK}
\textbf{Cluttered-Scene Manipulation: }Many works in task and motion planning (TAMP) \cite{pick_to_place4,pick_to_place5,Synergistic-Task-and-Motion-Planning,Tree-Search-based-Task-and-Motion-Planning} have achieved impressive results in solving cluttered-scene manipulation tasks but typically require poses of objects and known objects models. In contrast, some studies have employed RL to learn policies directly from visual inputs. Zeng et al.\cite{VPG} proposed a Visual Pushing Grasping (VPG) method that models push and grasp primitives by Fully Convolutional Networks (FCNs) \cite{FCN}. Babar et al. \cite{Prehensile_and_Non-Prehensile} and Tang et al. \cite{Selective_Object_Rearrangement} extended this approach in various ways. However, these methods \cite{Prehensile_and_Non-Prehensile,Selective_Object_Rearrangement} are limited by either hard-code placement primitives or the necessity for continual obstacle removal through picking and placing. Building on VPG, Kumra et al. \cite{Romanet} introduced the Robotic Manipulation Network (RoManNet). It adopted three FCN variants to model pushing, picking, and placing and employed TGP and LAE to augment the reward and exploration strategy. However, these VPG-based methods \cite{VPG,Prehensile_and_Non-Prehensile,Selective_Object_Rearrangement,Romanet} employ an \textit{implicit greedy policy} for selecting primitives, i.e., selecting the primitive and pixel with the highest Q value across all dense Q maps. Since each primitive network only uses the experience of executing its corresponding primitive for updates, the Q value within each primitive cannot capture potential rewards from other primitives. This hinders the policy's long-term planning for primitives. In this paper, we propose a hierarchical policy that explicitly adopts a high-level policy to select primitives, facilitating planning across multiple primitives.

\textbf{Transporter for Visual Manipulation: }
The recently proposed Transporter\cite{Transporter} has gained broad interest owing to its sample efficiency and the ability to complete multi-step manipulation tasks. It exploits the spatial consistency of orthographic images in tabletop manipulation tasks, modeling the spatial displacement of objects as a template-matching problem. Three FCNs are employed to generate the pick and place affordance maps. Various studies have modified or extended this method in different dimensions\cite{CLIPort,Transporter-1,Transporter-2,Transporter-3,Transporter-4}. Among them, CLIPort \cite{CLIPort} incorporates language instructions into the policy by using three separate encoders for RGB images, RGB-D images, and language instructions. It introduces a two-stream architecture to process different modalities. Inspired by CLIPort and recent advances in semantic segmentation\cite{two-stream}, we adopt a two-stream architecture to process RGB and depth channels. We also use skip connections to incorporate multi-scale information and late fusion to fuse features from two streams.

\textbf{Hierarchical Policy for Robot Manipulation:}
Inspired by the hierarchical learning present in human and animal cognition\cite{Intelligent-problem-solving,cognition1,cognition2}, many studies have employed hierarchical policies, consisting of a high-level policy and options (skills), to address long-horizon manipulation tasks \cite{RLBC,MAPLE,hierachical1,hierachical2,hierachical3}. When using HRL to train the high-level policy and options simultaneously, \textit{non-stationarity} is introduced to the high-level policy as the low-level policy is constantly changing\cite{non-stationary1}. Recent works have proposed various solutions to this issue, such as reward shaping \cite{MAPLE}, transition modification \cite{non-stationary2,non-stationary3}, or independent training\cite{RLBC,hierachical1,hierachical2}. However, both reward shaping and transition modification require supplemental information like option feasibility\cite{MAPLE} or specific goals\cite{non-stationary2,non-stationary3}. Moreover, training high-level policy and options independently will increase the required data and reduce the sample efficiency of the algorithm. In contrast, we propose a heuristic method, TSUS, which controls the update process of the high-level policy to alleviate the non-stationary transition problem without requiring additional information.

\section{PROBLEM FORMULATION}
We consider the problem of efficiently learning long-horizon manipulation in densely cluttered scenes. We introduce a repository of predefined semantic meaningful parameterized action primitives $\mathcal{R}=\{p^1,p^2,\dots,p^k\}$. Each primitive is a temporal abstraction of a sequence of basic actions $(b_1,b_2,\dots,b_t)$ and is designed to perform a family of subtasks. In our formulation, the basic action is the pose of the end-effector $b\in \mathbb{R}^7$ at the next timestep executed by the robot's built-in planner. The parameters of the primitive specify the exact basic action sequences. By sequentially selecting primitives, the policy can accomplish long-horizon tasks and adapt to variations within them.


We formulate the problem as a Parameterized Action Markov Decision Problem (PAMDP)\cite{POMDP}: at any given state $s_t \in \mathcal{S}$ at time step $t$, an agent receives an observation $o_t\in \mathcal{O}$ and executes an action $a_t=(a^H_t,a^L_t), a_t\in \mathcal{A}$ which is produced by the policy $\pi(a_t|s_t)$ of the agent. $a^H_t$ is the high-level action, $a^L_t$ is the low-level action. After executing the action, the environment will transit to a new state $s_{t+1}$ and return an immediate reward $r_t$ according to the reward function $R_{a_t}(o_t,o_{t+1})$. The policy aims to maximize the expected sum of discounted reward $R_t=\sum_{i=t}^{+\infty}\gamma_ir_i$. 


\begin{figure*}[t]
\centering
  \vspace{0.5em} 
  \includegraphics[width=\linewidth,scale=1.00]{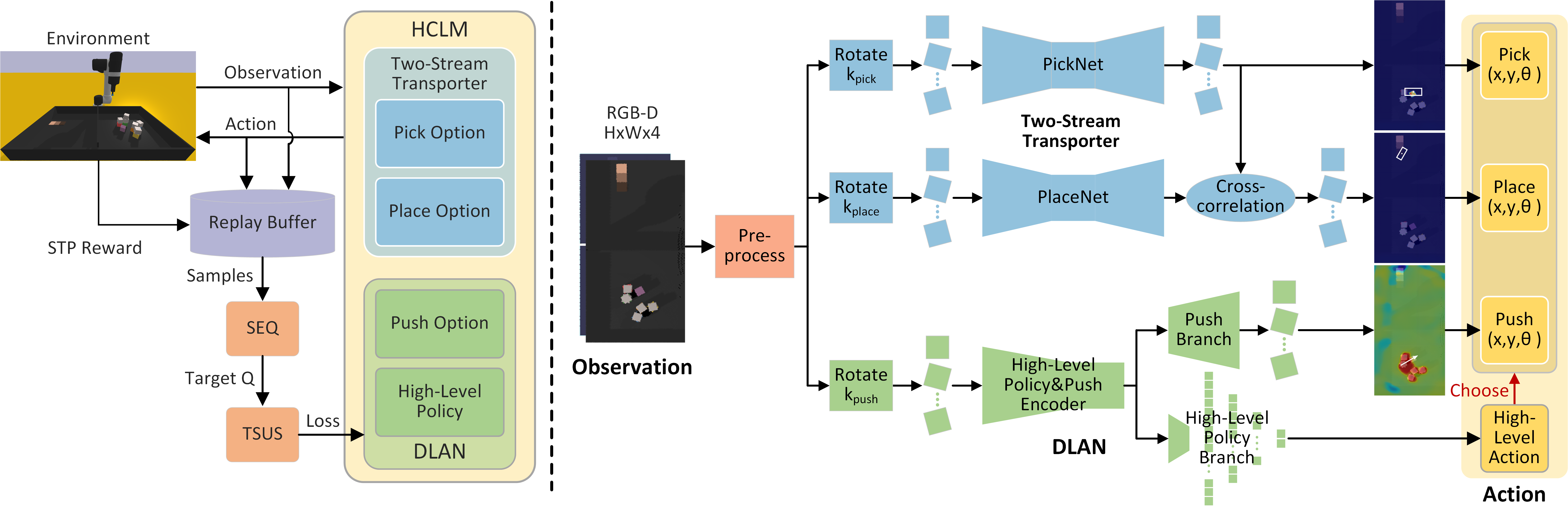}
  \vspace{-2.0em} 
  \caption{\textbf{Left:} After training the two-stream Transporter by BC, we freeze its parameters and employ HRL to train DLAN. During HRL, the HCLM policy generates actions based on observations and receives STP rewards from the environment. The interaction data will be stored in a replay buffer. We draw samples from the buffer and propose SEQ to calculate extended target Q values. Then, we propose TSUS to control the loss calculation of DLAN.
   \textbf{Right:} Architecture of the proposed HCLM policy, with a high-level policy that chooses primitive(s) to execute and three options that output primitive parameters. The pre-processed (resized, normalized) observations are rotated to different angles to model different primitive directions.
  }
  \label{fig:Policy}
\vspace{-1.7em} 
\end{figure*}

\section{METHODOLOGY}
In this study, we introduce three parameterized action primitives: push, pick, and place. To coordinate these primitives, we propose a hierarchical manipulation policy, HCLM, consisting of a high-level policy and three options. 
Specifically, we leverage a two-stream Transporter trained by BC to model the pick and place options and DLAN trained by HRL to model the high-level policy and push option. Given a current observation, the high-level policy outputs high-level actions to select primitive(s) for execution, while the options produce low-level actions to derive primitive parameters. During HRL, we augment Q-value updates for the push option via SEQ and propose TSUS to alleviate the non-stationary transition problem in updating the high-level policy. See Fig. \ref{fig:Policy} for an illustration.

\vspace{-0.5em}
\subsection{Observation Space and Action Space}
We model the visual observation $o_t \in \mathbb{R}^{H\times W \times 4}$ as an orthographic RGB-D image perpendicular to the tabletop. Subsequently, we parameterize the action space into two components. The first component is the high-level action at time $t$ , $a^H_t \in \{ push, pick\&place \} $, where $\{push\}$ denotes executing the push primitive, and $\{pick\&place\}$ denotes sequentially executing the pick and place primitives. The second component is the low-level action $a^L_t = (x,y,\theta)$, where $x,y$ specifies the position parameters in the image coordinate system, and $\theta$ defines the angle parameter. We extract primitive heights from the depth image.


In this paper, we use \textbf{u} to denote "push", \textbf{i} for "pick", and \textbf{l} for "place". The definitions for the parameterized action primitives and their parameters are as follows:
\begin{itemize}
    \item Pick action:
$a^{i} = (x,y,\theta)$ represents the position and angle for a top-down picking.
    \item Place action:
$a^{l} = (x,y,\theta)$ represents the position and angle for a top-down placing. Notably, a place action immediately follows a successful pick action.
    \item Push action:
$a^{u} = (x,y,\theta)$ represents the starting position and angle for a \SI{15}{cm} horizontal push. 
\end{itemize}


\begin{figure*}[t]
\centering
  \vspace{0.4em} 
  \includegraphics[width=\linewidth,scale=1.00]{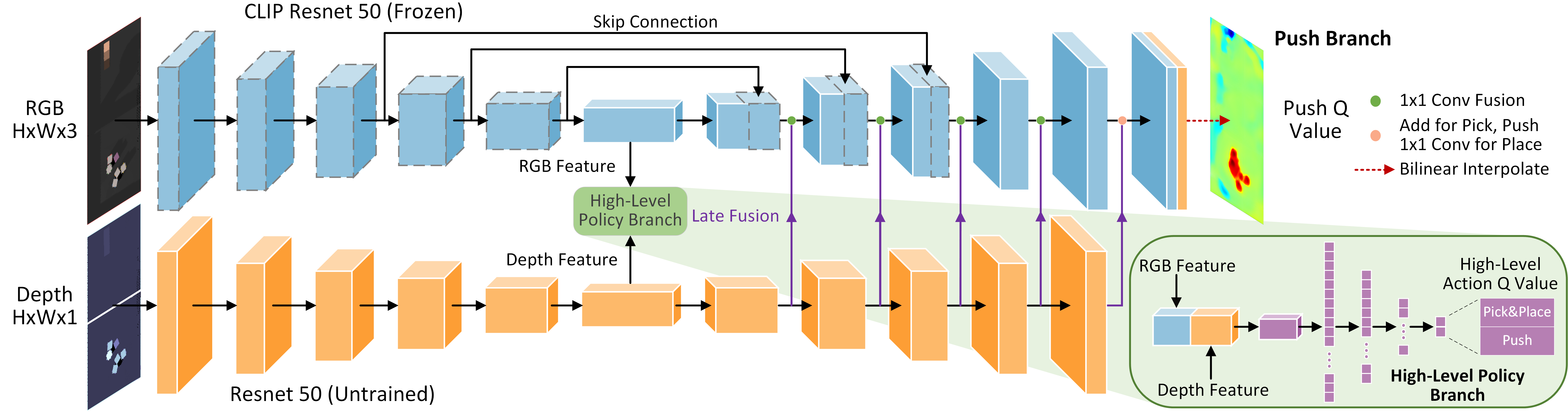}
\vspace{-1.7em} 
  \caption{\textbf{Dual-Level Action Network.} We implement a two-stream architecture to process RGB-D observation. Moreover, we incorporate two output branches to separately generate a dense push Q map and high-level action Q values. We employ a frozen CLIP Resnet50 to encode RGB images. In the RGB decoder, we introduce features from the encoder through skip connections. Besides, we utilize late fusion to integrate RGB and depth features.}
  \label{fig:Network}
\vspace{-1.7em} 
\end{figure*}
\vspace{-0.5em}
\subsection{High-Level Policy and Push Option}
In order to reduce model parameters and enhance training efficiency, we design a Dual-Level Action Network (DLAN) to model the high-level policy and push option concurrently. We employ two streams to process the RGB and depth channels separately and two branches to output the high-level and push action, respectively. Given an observation $o_t$, the push direction is modeled by rotating $o_t$ to $k_{push}$ angles. The push branch outputs a dense pixel-wise push Q map $Q_{push} \in \mathbb{R}^{H\times W \times k_{push}}$, where each pixel represents the Q-value estimate for a push at that location and angle. The push action is determined as follows: 
\begin{gather}
a^{u} = (x,y,\theta) = \mathop{\mathrm{argmax}}\limits_{(x,y,\theta)}{Q_{u}((x,y,\theta)|o_t)}.
\end{gather}
We use $k_{push}=12$ in this work. For the high-level policy branch, we only use the output from the non-rotated observation to predict the Q-values for high-level actions. The high-level action is then given by: 
\begin{gather}
a^H = \mathop{\mathrm{argmax}}\limits_{a^H}{Q_{H}(a^H|o_t)}.
\end{gather}
We employ HRL to train the high-level policy and push options simultaneously. 


In terms of network structure, the push branch and the high-level policy branch share a two-stream encoder to extract features from RGB-D observations. For the push branch, following methods in CLIPort\cite{CLIPort} and semantic segmentation\cite{two-stream}, we utilize skip connections to introduce features of varying scales from the encoder, thereby fusing low-level details with high-level semantic information. Additionally, we employ late fusion to merge features from two modalities. For the high-level policy branch, features from both streams are merged through concatenation, which are then processed through a convolution block and linear layers to predict the Q-values for high-level actions. See Fig. \ref{fig:Network} for an overview of our network architecture. 

\vspace{-0.1em}
\subsection{Pick and Place Options}
For the pick and place options, we improve upon the original Transporter by extending FCNs to two-stream architectures. This extended network architecture is identical to the two-stream network in DLAN when removing the high-level policy branch of DLAN. Analogous to the push option, we model the directions for pick and place by rotating the observation $o_t$ to $k_{pick}$ and $k_{place}$ angles, respectively. In this study, we set $k_{pick}=1$ and $k_{place}=36$. For a comprehensive explanation of how the Transporter derives the pick and place actions, we refer readers to the original work \cite{Transporter}. However, as noted in that work, Transporter faces challenges from partial visibility and occlusion \cite{Transporter}. When the target object is occluded by other objects, the high-level policy can select the push option to remove obstructions. So that the pick and place options would only need to operate unoccluded objects, addressing the challenges above.


\subsection{Stepwise Task Progression (STP) Reward}
As manipulation tasks can be divided into several subtasks, we design a reward function named the Stepwise Task Progression (STP) reward to reward actions that complete the current subtask and penalize actions that cause the task progress to regress. 
Specifically, our reward function is defined as follows:
\begin{gather}
R_{t}= 
\begin{cases}
\Delta_{task\_progress} , \; \textrm{if}\; \Delta_{task\_progress} < 0\\
\mathcal{W}(a_t^H)\mathcal{I}(a_t,s_{t+1}) , \; \textrm{otherwise}
\end{cases}.
\end{gather}
Here, $\Delta_{task\_progress}$ denotes the change in task progress after executing $a_t$. The reward weight, $\mathcal{W}(a_t^H)$, is 0.75 for a successful ${push}$ action and 1 for a successful ${pick\&place}$ action. $\mathcal{I}(a_t,s_{t+1})$ is an indicator function that is 1 when action $a_t$ is successful and 0 otherwise. In our experiments, a successful ${push}$ action is defined as lowering the height of at least one object, and a successful ${pick\&place}$ action is defined as successfully picking and placing a target object in the next subtask.


\begin{figure}[tb]
\centering
  \includegraphics[width=0.65\linewidth,scale=1.00]{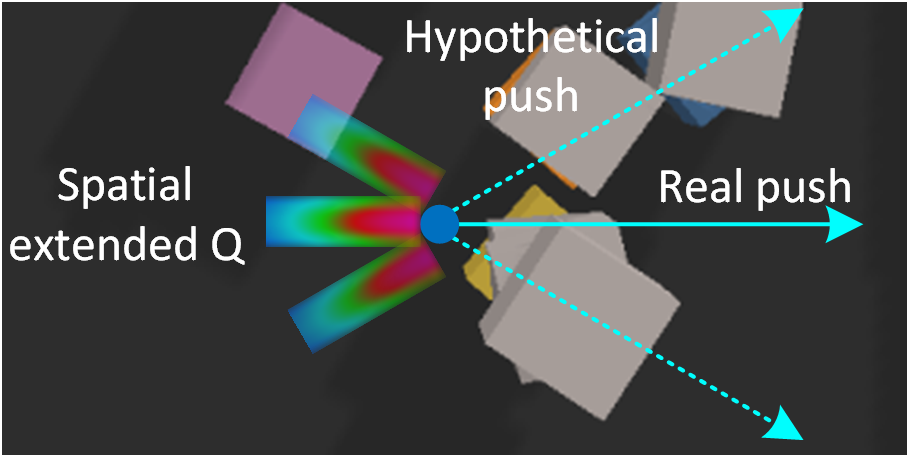}
    \vspace{-0.7em} 
  \caption{\textbf{Illustration of SEQ.} The solid line represents the real push, and the dashed line represents the hypothetical push. 
  Given the push action's inherent directionality, we employ anisotropic Gaussian distribution to spread the Q-value of the push action from a single pixel to a pixel area. Further, the Q-value of the real push can be expanded to incorporate two nearby hypothetical pushes, with a discount factor indicating the performance diminution of the hypothetical push.}
  \label{fig:Gaussian}
\vspace{-1.7em} 
\end{figure}

\subsection{Spatially Extended Q-Update (SEQ)}
For the push option, each pixel in the push Q map corresponds to a push action. When training the push option, VPG\cite{VPG} only computes the loss for the pixel corresponding to the taken push action. TPG \cite{Romanet}, additionally, calculates the loss for all pixels within a square region centered around the selected pixel. However, given the inherent directionality of the push action, its outcome can propagate oppositely over a certain region, implying that outcomes within this region may be largely consistent. We use an anisotropic Gaussian distribution to depict this consistency, extending the Q value from a single pixel to a region. Furthermore, as the push direction is discretized into $k_{push}$ angles, two adjacent push directions may yield similar outcomes. We use a degradation factor to account for the performance degradation for adjacent push angles. We refer to this updating method as the Spatially Extended Q-update (SEQ). See Fig. \ref{fig:Gaussian} for an illustration of SEQ. It is defined as follows:

\vspace{-0.2em} 
\begin{equation}
    \begin{aligned}
        \mathbf{Filter} = \left[\dfrac{1}{2\pi \sigma_x \sigma_y}\exp \left(-\dfrac{x^2}{2\sigma_x^2}-\dfrac{y^2}{2\sigma_y^2}\right)\right],
    \end{aligned}
\end{equation}
\vspace{-1.3em} 
\begin{gather}
	\mathbf{Y}_t^{u} = R_t \cdot \mathbf{Filter}
	+ \gamma \eta Q_{u}(o_{t^{\prime}}, \mathop{\mathrm{argmax}}\limits_{a_{t^{\prime}}}{Q_{u}(o_{t^{\prime}},a_{t^{\prime}})}),
\end{gather}
\vspace{-1.3em} 
\begin{gather}
\boldsymbol{\delta}_t^{u} = \Big[\mathbf{Q}^{prev}_u,\mathbf{Q}_u,\mathbf{Q}^{next}_u\Big]-\Big[\kappa \mathbf{Y}_t^{u},\mathbf{Y}_t^{u},\kappa \mathbf{Y}_t^{u}\Big],
\end{gather}
where $x,y \in \mathbb{Z},\, -k_x\leq x\leq0,\, |y| \leq k_y$. We define $k_x$ and $2k_y+1$ as the width and height of the region. $\gamma$ is the discount factor. $\eta$ is the reward discriminant factor. When $R_{a_t}(s_t,s_{t+1})>0$, $\eta=1$; otherwise, $\eta=0$. $\eta$ ensures that the policy prioritizes current task progress and considers subsequent Q values only if the current subtask is completed. $\kappa$ is the degradation factor used to capture the decrease in target Q value for adjacent push angles. $Q_{prev}$ and $Q_{next}$ are the Q-value of the previous and next push angles.


\begin{figure}[t]
\vspace{0.4em} 
\centering
  \includegraphics[width=\linewidth,scale=1.00]{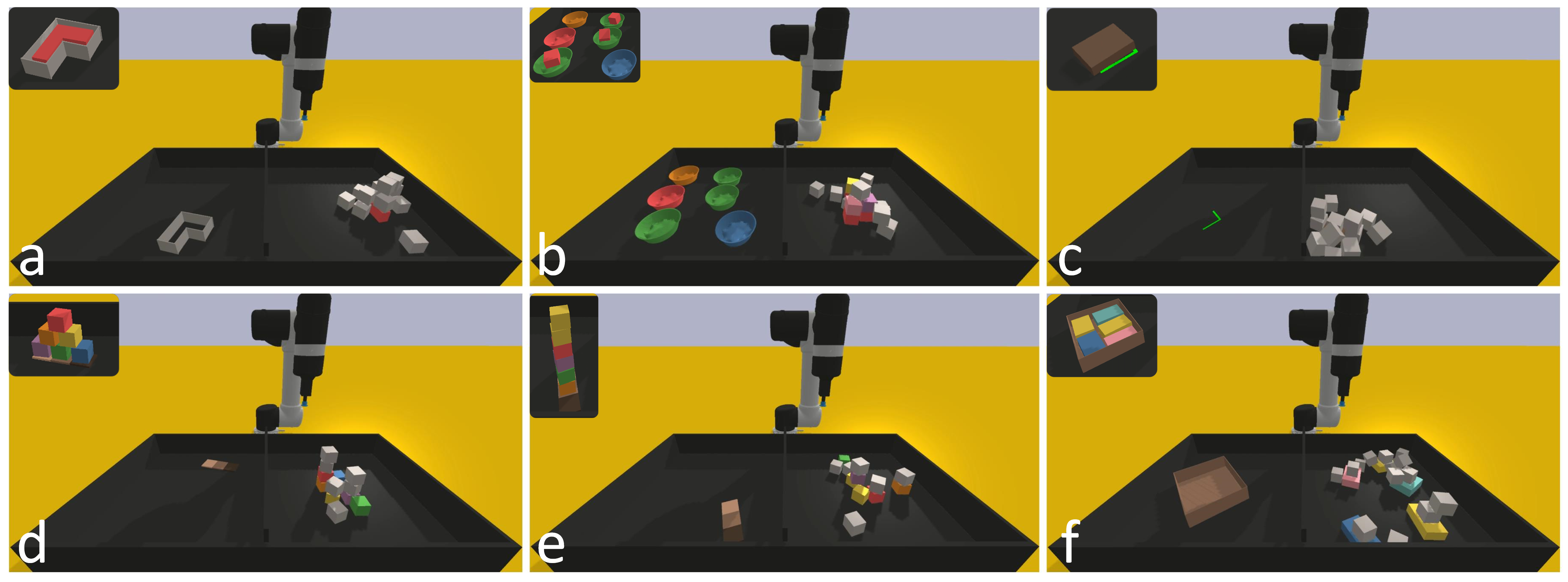}
 \vspace{-2.0em} 
  \caption{\textbf{Simulated Environments.} We perform evaluations on six tasks in ClutteredRavens. \textbf{Tasks:} (a) cluttered-block-insertion, (b) cluttered-place-red-in-green, (c) cluttered-align-box-corner, (d) cluttered-stack-block-pyramid, (e) cluttered-block-stacking, (f) cluttered-packing-boxes.}
  \label{fig:simulated environment}
\vspace{-1.9em} 
\end{figure}

\subsection{Hybrid Training Method and Two-Stage Update Scheme}
For the pick and place options, training from scratch using reinforcement learning is highly challenging due to the large exploration space and sparse reward function\cite{RLBC}. Hence, we employ behavior cloning to train it on the demonstration dataset  $ \mathcal{D}=\left\{ (o_1,a_1), (o_2,a_2), \ldots, (o_n,a_n) \right\} $. During training, we randomly sample a tuple from the dataset and expand the action into a one-hot action label $Y:\mathbb{R}^{H\times W \times k}$. We use cross-entropy loss for training: $\mathcal{L}=-\mathbb{E}_{Y}\left[\log(\textrm{softmax}(Q)\right]$.


After training the pick and place options, we select the best-performing checkpoint in the non-cluttered environment and freeze its parameters. Then, we employ HRL with a Prioritized Replay Buffer\cite{PER} to simultaneously train the high-level policy and push option. We adopt $\epsilon$-greedy exploration strategies for the high-level policy and the push option. For the high-level policy, $\epsilon=1$. For the push option, $\epsilon$ is initially set to 0.5 and anneal to 0.1 during training. 


For the push option, our goal of the learning process is to progressively minimize the temporal difference error $\boldsymbol{\delta}_t^{u}$. We adopt Huber loss for training: $\mathcal{L}_t^{u}= \mathcal{X}(a_t^H)\cdot\mathcal{H}(\boldsymbol{\delta}_t^{u})$. Here, $\mathcal{H}$ is Huber loss, and $\mathcal{X}$ is the action discriminator. When $a_t^H={push}$, $\mathcal{X}=1$; otherwise, $\mathcal{X}=0$, i.e., we only update with the data of performing push action. 


As discussed in section \uppercase\expandafter{\romannumeral2}, when simultaneously training the high-level policy and push option, the updates of the high-level policy confront the non-stationary transition problem. We propose a heuristic method, Two-Stage Update Scheme (TSUS), to mitigate this problem. The intuition is that, during the early stage of training, as the output of the push option is unreliable, we only use the successful push experiences to update the high-level policy. Once the training process surpasses a threshold $\tau$, we assume that the push option can produce reasonable actions. Hence, we use all non-random push experiences to update the high-level policy. For experiences from random pushes, we only use the successful push experiences for updates. The specific definition of TSUS is as follows:

\vspace{-1.0em}
\begin{gather}
\begin{split}
    y_t^{H} = R_{t} 
    + \gamma \eta Q_H(o_{t^{\prime}},\mathop{\mathrm{argmax}}\limits_{a_{t^{\prime}}}{Q_{H}(o_{t^{\prime}},a_{t^{\prime}})}),
\end{split}\\
\delta_t^{H} = Q_{H}(o_t,a_t)-y_t^{H},\\
\mathcal{L}_t^{H}= 
\begin{cases}
\mathcal{H}(\delta_t^{H}) , \; \textrm{for}\; a_t^H=\{pick\&place\}\\
\mathcal{P}(a_t,r_t,n)\cdot\mathcal{H}(\delta_t^{H}) , \; \textrm{for}\; a_t^H=\{push\}
\end{cases},\\
\mathcal{P}(a_t,r_t,n)= 
\begin{cases}
u(r_t) , \; \textrm{for}\; n<\tau\\
u(r_t) \lor \Psi(a_t) , \; \textrm{otherwise}
\end{cases}.
\end{gather}
Here, $n$ is the current training epoch. $u(r_t)$ is a step function, which equals to $1$ when $r_t>0$ and $0$ otherwise. $\Psi(a_t)$ is a discriminant function that returns $0$ for a random push, and $1$ otherwise. $\tau$ denotes the update changing threshold.


\section{EXPERIMENTS}
We conduct our experiments to answer three questions: 1) How well does our method perform on diverse long-horizon manipulation tasks in densely cluttered scenes compared to baselines? 2) Is our method capable of generalizing to more cluttered environments? 3) What influence does each component of our method possess on overall performance?



\subsection{Experiment Setup and Tasks Design}



\begin{table}[t]
\vspace{0.4em} 
  \setlength\tabcolsep{2.3pt}
  \centering
    \caption{Tasks and Their Attributes}
    \label{tab:Tasks and Attributes}
  \vspace{-0.8em}
  \begin{tabular}{@{}lcccc}
  \toprule
  & precise & multimodal & pick\&place & unseen \\
  Task & placing & placing & sequencing & objects \\
  \midrule
  cluttered-block-insertion    & \cmark & \xmark & \xmark & \xmark \\
  cluttered-place-red-in-green                                 & \xmark & \cmark & \xmark & \xmark \\
  cluttered-align-box-corner    & \cmark & \cmark & \xmark & \cmark \\
  cluttered-stack-block-pyramid                                   & \cmark & \cmark & \cmark & \xmark \\
  cluttered-block-stacking                        & \cmark & \cmark & \xmark & \xmark \\
  cluttered-packing-boxes                        & \cmark & \cmark & \cmark & \cmark \\
  \bottomrule
  \end{tabular}
  \vspace{-1.8em}
\end{table}

Based on the Ravens benchmark\cite{Transporter} set in PyBullet\cite{pybullet}, we propose the ClutteredRavens benchmark consisting of six diverse long-horizon manipulation tasks in densely cluttered environments, showcased in Fig. \ref{fig:simulated environment}. Each task and its attributes are detailed in Table. \ref{tab:Tasks and Attributes}. We employ a Universal Robot UR5e with a suction gripper for manipulation and acquire observations in the same way as Ravens. Static objects (base, bowl, etc.) are randomly generated on the left side of the workspace, while target objects are randomly generated on the right. To create densely cluttered scenes, task-irrelevant additional blocks (grey) are dropped from above the target objects. An oracle is provided for each task to offer expert demonstrations for pick and place.


\begin{table*}[t]
\vspace{0.4em} 
\begin{center}
\setlength\tabcolsep{4pt}
\caption{The performance of HCLM and baselines on tasks (Mean success rate \%, Average episode length).}
\label{tab:HCLM and baselines}
\begin{tabular}{l cc cc cc}
\toprule
 & \multicolumn{2}{c}{cluttered-block-insertion} &  \multicolumn{2}{c}{cluttered-place-red-in-green} & \multicolumn{2}{c}{cluttered-align-box-corner} \\

\cmidrule(lr){2-3} \cmidrule(lr){4-5} \cmidrule(lr){6-7} 

Method  & Success Rate & Episode Length & Success Rate & Episode Length & Success Rate & Episode Length \\

\midrule

Pick\&Place-only &
54 & 4.73 & 42 & 8.16 & 64 & 4.49 \\

Pick\&Place+Random push &
36 & 5.42 & 42 & 9.54 & 51 & 6.24 \\

RoManNet & 5 & 6.70 & 71 & 7.29 & 0 & 7.26 \\

RoManNet-demo & 56 & 3.39 & 90 & 5.12 &
89 & \textbf{4.08} \\

\textbf{HCLM (Ours)} &
\textbf{97} & \textbf{3.28} & \textbf{97} & \textbf{3.13} &
\textbf{95} & 4.52 \\

\midrule
 & \multicolumn{2}{c}{cluttered-stack-block-pyramid} &  \multicolumn{2}{c}{cluttered-block-stacking} & \multicolumn{2}{c}{cluttered-packing-boxes} \\

\cmidrule(lr){2-3} \cmidrule(lr){4-5} \cmidrule(lr){6-7} 

 & Success Rate & Episode Length & Success Rate & Episode Length & Success Rate & Episode Length \\

\midrule

Pick\&Place-only & 19 & 18.02 & 31 & 16.07 & 43 & 19.37 \\

Pick\&Place+Random push & 21 & 18.80 & 26 & 18.44 &
35 & 23.26 \\

RoManNet & 0 & 18.69 & 0 & 17.75 & 9 & 28.46 \\

RoManNet-demo & 24 & 17.50 & 31 & 17.19 & 63 & \textbf{12.34} \\

\textbf{HCLM (Ours)} & \textbf{87} & \textbf{10.95}  & \textbf{91} & \textbf{8.80} & \textbf{77} & 16.31 \\

\bottomrule
\end{tabular}
\end{center}
\vspace{-1.9em} 
\end{table*}

\subsection{Evaluation Metrics}
We employ the following metrics to evaluate performance:

\begin{itemize}
    \item Task success rate:
    the percentage of tasks completed successfully within the task's max steps. 
    \item Average episode length:
    the average step count across all episodes. Lower values indicate higher efficiency.
\end{itemize}

All models are trained on the same train seed. After each epoch, the model is validated for 100 episodes on the validation seed. We save the parameters that yield the highest success rate to date. 
The best-validated model is tested for 100 episodes on the test seed to obtain final performance. 


\subsection{Baselines}
\textbf{Pick\&Place-only} includes only the pick and place options, modeled by the two-stream Transporter, without the push option and high-level policy. \textbf{Pick\&Place+Random push}, on top of Pick\&Place-only, uses random pushing as the push option and an alternating policy as the high-level policy. The alternating policy alternately outputs ${pick\&place}$ and ${push}$ as high-level action. \textbf{RoManNet}, as mentioned in Related Work, employs three independent GR-ConvNet\cite{GR-ConvNet} to model the pick, place, and push actions rather than a hierarchical policy structure. It augments rewards by TPG and reduces unnecessary exploration by LAE. The policy is entirely trained via RL. \textbf{RoManNet-demo} uses BC to train the pick and place network on the same dataset as the two-stream Transporter. It also adopts an alternating policy as the high-level policy to train the push network.


\subsection{Results}

\textbf{The performance of HCLM and baselines on ClutteredRavens}: 
As shown in Table \ref{tab:HCLM and baselines}, HCLM outperforms the baselines in mean success rate across all tasks and achieves the lowest mean episode length in four tasks. First, we see that the inclusion of push primitive allows HCLM to considerably outperform the Pick\&Place-only baseline. Next, we find that the Pick\&Place+Random push performs worse than Pick\&Place-only, indicating that random pushes without planning can be counterproductive. While the RoManNet manages to solve cluttered-place-red-in-green, it struggles with tasks requiring precise placing. In contrast, the RoManNet-demo shows a significant performance improvement across all tasks. This highlights the substantial exploration challenge presented by the sparse rewards of precise placing in RL. Collectively, these results emphasize that by leveraging HRL and BC to train a hierarchical policy, we can effectively coordinate the push, pick, and place primitives to address a wide range of cluttered-scene long-horizon manipulation tasks.



\begin{table}
\begin{center}
\caption{Task success rate on increasing additional blocks. (Mean \%) vs number of additional blocks}
\label{tab:increasing additional}
\begin{tabular}{l c c c c c c}
\toprule
Method & 6  & 8  & 10  & 12  & 14  & 16    \\
\midrule
Pick\&Place-only  & 19 & 23 & 21 & 13 & 22 & 20 \\
Pick\&Place+Random push  & 21 & 14 & 17 & 14 & 18 & 24 \\
RoManNet  & 0 & 0 & 0 & 0 & 0 & 0 \\
RoManNet-demo  & 24 & 33 & 31 & 30 & 22 & 34 \\
\midrule
\textbf{HCLM (Ours)} & \textbf{87} & \textbf{83} & \textbf{81} & \textbf{76} & \textbf{76} & \textbf{70} \\
\bottomrule
\end{tabular}
\end{center}
\vspace{-2.2em} 
\end{table}

\textbf{Generalization to increasing additional blocks}: 
To verify the policy's generalizability to more cluttered environments, we progressively add more additional blocks in the cluttered-stack-block-pyramid task. Notably, all models are trained in environments with only six additional blocks. As shown in Table \ref{tab:increasing additional}, HCLM consistently outperforms other baselines, maintaining the 70\% success rate even with 16 additional blocks. These results demonstrate HCLM's robust generalization across diverse clutter situations.


\begin{table}
\begin{center}
\caption{Ablation studies. (Mean success rate \%, Average episode length).}
\label{tab:ablation studies}
\begin{tabular}{l c c}
\toprule
Method & Success Rate  & Episode Length  \\
\midrule
HCLM with original Transporter   & 71 & 12.87  \\
HCLM w/o hierarchical policy   & 0 & 20 \\
HCLM w/o TSUS   & 52 & 14.28 \\
HCLM w/o SEQ   & 79 & 11.98 \\
\midrule
\textbf{HCLM (Ours)} & \textbf{87} & \textbf{10.95} \\
\bottomrule
\end{tabular}
\end{center}
\vspace{-2.2em} 
\end{table}

\textbf{Ablation studies}: 
We conduct ablation studies to validate the effectiveness of each module in HCLM. We compare HCLM with: 1) replacing the pick and place options with the original Transporter (\textbf{HCLM with original Transporter}); 2) without the hierarchical policy structure, same as RoManNet (\textbf{HCLM w/o hierarchical policy}); 3) without the Two-Stage Update Scheme (\textbf{HCLM w/o TSUS}); 4) without the Spatially Extended Q-update (\textbf{HCLM w/o SEQ}). As shown in Table \ref{tab:ablation studies}, removing any module degrades the policy's performance. Notably, the absence of the hierarchical policy structure makes task completion infeasible. We hypothesize that this is due to inconsistent optimization goals between BC (for pick and place ) and RL (for push), which leads to inconsistent value outputs. This makes the implicit greedy strategy infeasible.

\section{CONCLUSION}

In this work, we propose a vision-based hierarchical manipulation policy, HCLM. By coordinating push, pick, and place parameterized primitives, it can efficiently perform diverse long-horizon manipulation tasks in densely cluttered scenes. Our experiments demonstrate that our method can generalize to more cluttered environments. Through ablation studies, we validate the effectiveness of each component of our method. Future studies can explore more flexible solutions to the non-stationary transition problem and extend this framework with more primitives.










\clearpage

\bibliographystyle{IEEEtran}
\balance
\bibliography{root}


\end{document}